\documentclass{article}
\usepackage{ijcai24}

\usepackage{times}
\usepackage{soul}
\usepackage{url}
\usepackage[hidelinks]{hyperref}
\usepackage[utf8]{inputenc}
\usepackage[small]{caption}
\usepackage{graphicx}
\usepackage{amsmath}
\usepackage{amsthm}
\usepackage{booktabs}
\usepackage{algorithm}
\usepackage{algorithmic}
\usepackage[switch]{lineno}
\usepackage{subfigure}
\usepackage{amssymb}
\usepackage{multirow}
\usepackage{threeparttable}
\usepackage{enumitem}
\usepackage{amsfonts}
\usepackage{float}
\newcommand{\model}[1]{FedGCS}
\newcommand\scalemath[2]{\scalebox{#1}{\mbox{\ensuremath{\displaystyle #2}}}}

\title{\textit{\model~}: A Generative Framework for Efficient Client Selection in Federated Learning via Gradient-based Optimization}

\newcommand*\samethanks[1][\value{footnote}]{\footnotemark[#1]}

\author{
Zhiyuan Ning$^{1,2}$\thanks{These authors contributed equally.}
\and
Chunlin Tian$^{4}$\samethanks[1]
\and
Meng Xiao$^{1,2}$
\and
Wei Fan$^{5}$
\and
Pengyang Wang$^{4}$
\and 
Li Li$^{4}$\thanks{Corresponding author.}
\and \\
Pengfei Wang$^{1,2}$\samethanks[2]
\And
Yuanchun Zhou$^{1,2,3}$\\
\affiliations
$^1$Computer Network Information Center, Chinese Academy of Sciences\\
$^2$University of Chinese Academy of Sciences 
$^3$Hangzhou Institute for Advanced Study, UCAS\\
$^4$Department of Computer and Information Science, IOTSC, University of Macau 
$^5$University of Oxford\\
\emails
ningzhiyuan@cnic.cn,
yc27402@um.edu.mo,
shaow@cnic.cn,
wei.fan@wrh.ox.ac.uk,\\
\{pywang, llili\}@um.edu.mo,
\{pfwang, zyc\}@cnic.cn
}

\begin{document}

\maketitle

\begin{abstract}
Federated Learning faces significant challenges in statistical and system heterogeneity, along with high energy consumption, necessitating efficient client selection strategies. 
Traditional approaches, including heuristic and learning-based methods, fall short of addressing these complexities holistically. 
In response, we propose \model~, a novel generative client selection framework that innovatively recasts the client selection process as a generative task. 
Drawing inspiration from the methodologies used in large language models, \model~ efficiently encodes abundant decision-making knowledge within a continuous representation space, enabling efficient gradient-based optimization to search for optimal client selection that will be finally output via generation. 
The framework comprises four steps: 
(1) automatic collection of diverse ``selection-score'' pair data using classical client selection methods; 
(2) training an encoder-evaluator-decoder framework on this data to construct a continuous representation space; 
(3) employing gradient-based optimization in this space for optimal client selection; 
(4) generating the final optimal client selection via using beam search for the well-trained decoder. 
\model~ outperforms traditional methods by being more comprehensive, generalizable, and efficient, simultaneously optimizing for model performance, latency, and energy consumption. 
The effectiveness of \model~ is proven through extensive experimental analyses.
\end{abstract}
\vspace{-2mm}
\section{Introduction}
Federated Learning (FL), as introduced in~\cite{fedavg}, enables multiple client devices to collaboratively train a shared model without exposing users' raw data, preserving privacy. 
Despite its advantages, the practical deployment of FL still faces various challenges: 
(1) statistical heterogeneity, \textit{i.e.}, data owned by different clients may come from different distributions, resulting in non-independent, identically distributed (Non-IID) training data, which can seriously affect the convergence and overall performance of the global model~\cite{hsieh2020non};
(2) system heterogeneity, \textit{i.e.}, clients participating in FL have different degrees of computational resources, communication capabilities and fault tolerance, which may result in stragglers that hinder the FL training process inflicting intolerable latency~\cite{bonawitz2019towards}; (3) the devices perform many compute-intensive data processing and model training, and the devices and the central server frequently exchange model parameters and gradients, which will lead to substantial energy consumption.
Thus, the inclusion of too many clients may lead to suboptimal performance, latency, and wasted energy. 
The selection of a ``good'' subset of clients as FL participants for each training round is critical to mitigating all three of these issues~\cite{oort}.

\begin{figure}[t]
    \centering
    \vspace{-2mm}
    \includegraphics[width =1.0\linewidth]{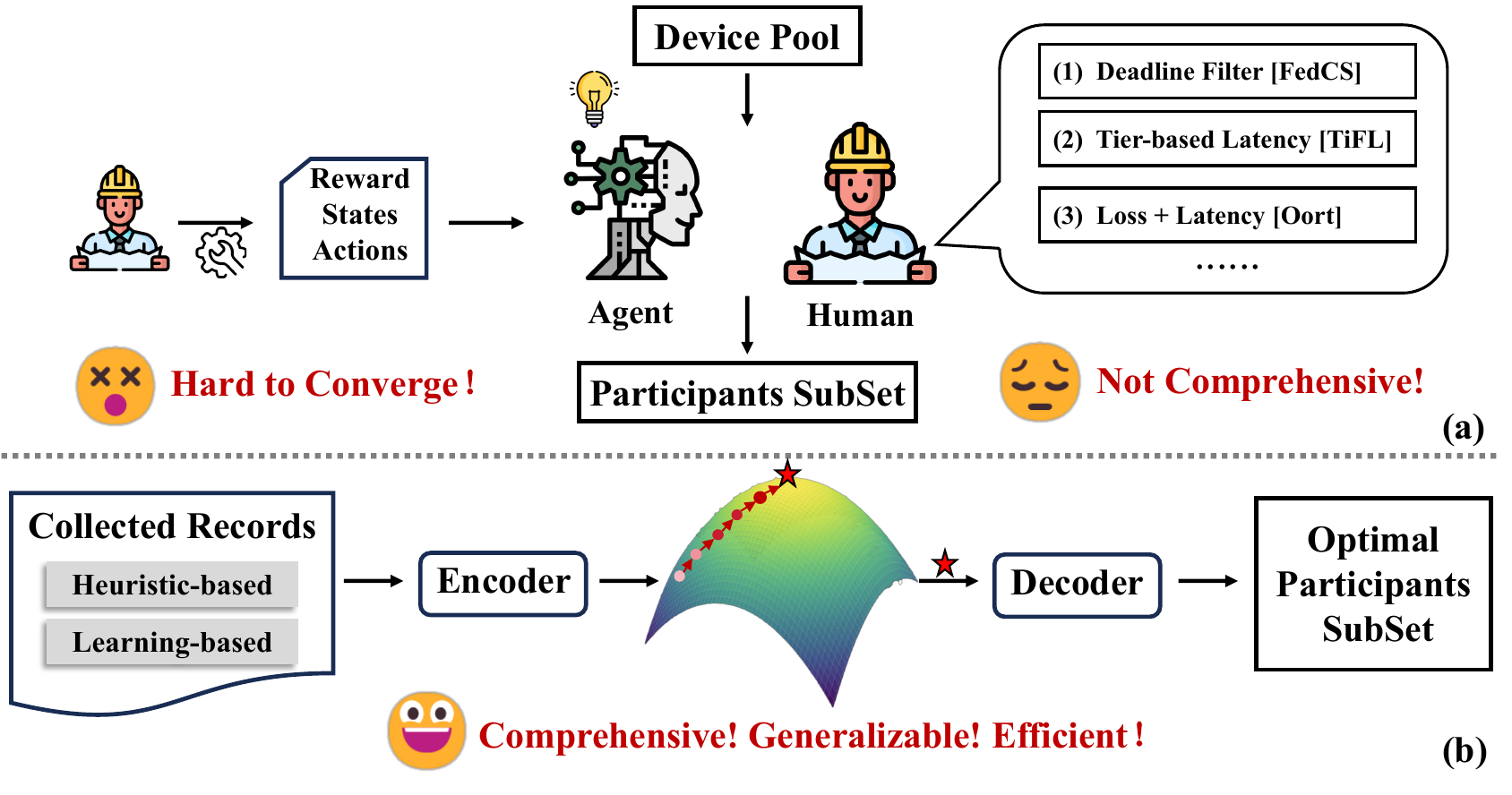}
    \vspace{-6mm}
    \caption{(a) Right: heuristic-based selection, which utilizes manually developed heuristic policies to select participating devices. (a) Left: learning-based selection, which employs RL to train continuously self-optimizing agents for optimal client selection. (b): \model~ formulates client selection as a generative task.}
    \label{fig:intro}
    \vspace{-4.5mm}
\end{figure}

Existing client selection methods only partially address the above three challenges and can be categorized into two types:
(1) \textit{heuristic-based selection}, as shown in Figure~\ref{fig:intro}(a) right, in which manually-engineered heuristic rules are used to choose participating devices to improve the model performance and training efficiency~\cite{tifl,oort,DBLP:conf/micro/TianLSW022}.
However, such methods are subjectively designed by human experts and are not comprehensive enough that consider only a limited number of optimization objectives and deployment scenarios.
As a result, when confronted with unfamiliar and complex scenarios, a great deal of domain expertise and manual tuning is often required, and the final performance may be suboptimal;
(2) \textit{learning-based selection}, as shown in Figure~\ref{fig:intro}(a) left, which explores the use of reinforcement learning (RL)~\cite{reinforcement} to train continuously self-optimizing agents for optimal client selection~\cite{Favor,fedmarl,DBLP:conf/iclr/TianSL23}.
However, the objectives designated by human experts for RL training still do not handle the evolving and complex real-world deployment scenarios well~\cite{kim2021autofl}. 
Moreover, RL is developed based on making decisions in massive discrete spaces and therefore has difficulty converging compared to solving continuous optimization problems~\cite{dulac2021challenges}.

Recently, large language models (LLMs) trained on large volumes of text data have shown impressive abilities to encode world knowledge into model parameters and solve a wide variety of natural language processing tasks via text generation~\cite{brown2020language,touvron2023llama}.
Such great success has driven various domains to migrate to generative models~\cite{ramesh2022hierarchical,gruver2023large}, where all the information and knowledge required for a task is encoded into the model parameters and the task is ultimately solved in a generative manner.
This is a general, flexible, comprehensive, and efficient modeling style that prompts us to consider: \textbf{\textit{can client selection in FL also be effectively addressed in a similar generative way?}}
To this end, we propose \textbf{\model~}, a novel \textbf{G}enerative \textbf{C}lient \textbf{S}election framework for \textbf{Fed}erated learning, that formulates the discrete client device selection task as a generative task.
Like LLMs, \model~ tries to encode the wide knowledge of discrete decisions (\textit{i.e.}, selection or deselection of clients) into a continuous representation space, where gradient-based optimization can be efficiently applied to find the optimal representation that will eventually be output as the discrete selection format via generation (as shown in Figure~\ref{fig:intro}(b)).
Specifically, \model~ includes four steps:
(1) utilizing classical client selection approaches, \textit{i.e.}, heuristic-based and learning-based selection, to automatically collect sufficient, diverse and high-quality ``selection-score'' pair data as training data for subsequent models;
(2) training an encoder-evaluator-decoder framework by simultaneously optimizing the sequence-to-sequence loss~\cite{sutskever2014sequence} and score estimation loss based on the collected pair data, and thereby obtaining a continuous representation space on which selection optimization will be performed;
(3) adopting gradient-based optimization in the continuous representation space to find the optimal client selection representation;
(4) applying the beam search strategy~\cite{freitag-al-onaizan-2017-beam} to the well-trained decoder to generate the optimal client selection based on the optimal representation.

\model~ has the following advantages over previous methods:
(1) \model~ encodes abundant experience of various different classical algorithms into the same continuous space to determine the final decision, and simultaneously considers the three metrics of model performance, latency and energy, making it more \textbf{\textit{comprehensive}} than the traditional methods;
(2) through data-driven learning, \model~ only needs the automatically collected ``selection-score'' pair data as input and avoids the overly complex and cumbersome manual interventions and tuning of previous methods, therefore \model~ is more flexible and \textbf{\textit{generalizable}};
(3) modeling the process of determining the optimal client selection as executing gradient-based optimization in a continuous space is more \textbf{\textit{efficient}} than performing a heuristic rule-based search in a large discrete space or training hard-to-converge RL agents.
Finally, we conduct extensive experiments and analyses to demonstrate the effectiveness and superiority of \model~.
% \vspace{-2mm}
\section{Related Works}
\begin{figure}[t]
    \centering
    \includegraphics[width=0.48\linewidth]{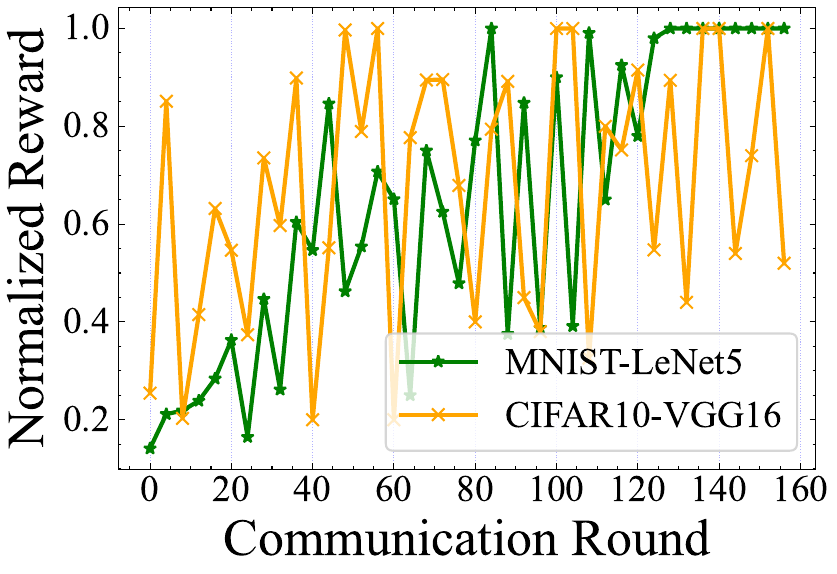}
    \includegraphics[width=0.48\linewidth]{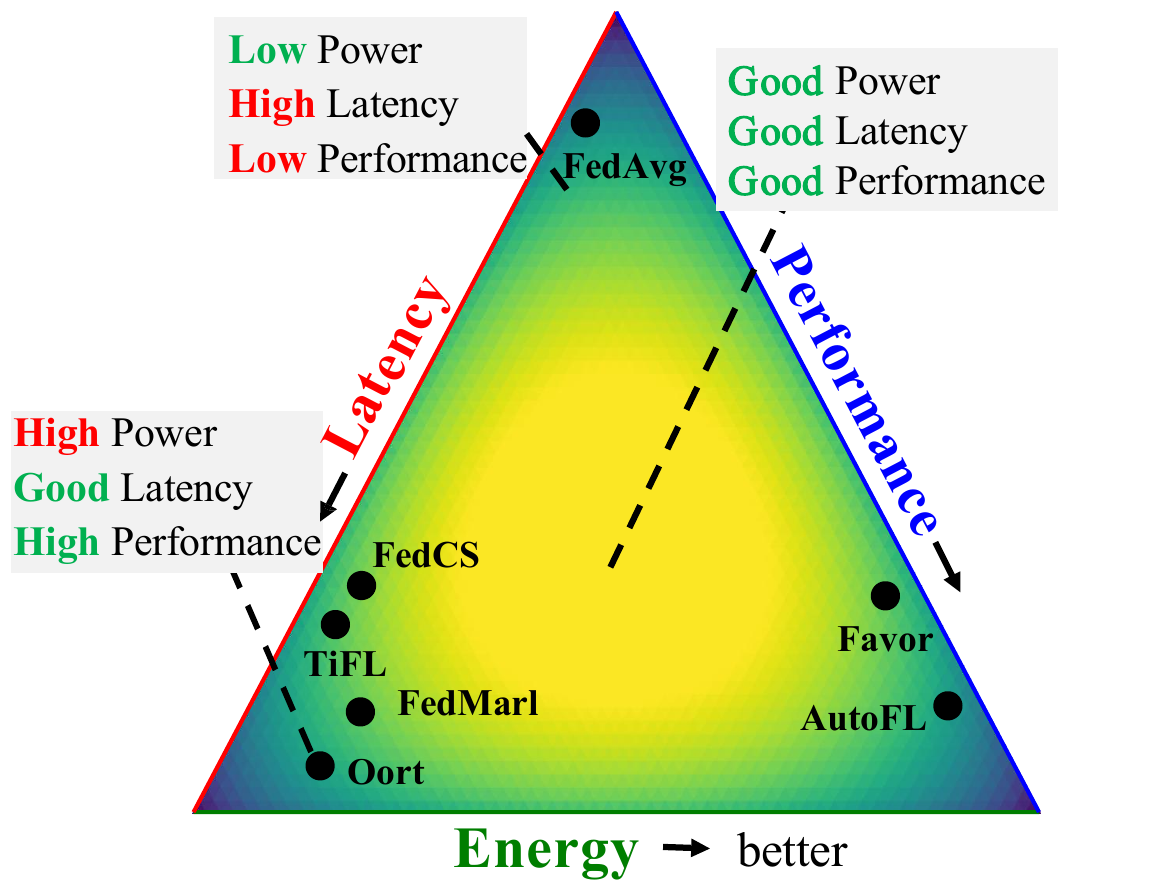}
    \vspace{-2mm}
    \caption{Left: reward convergence curve of Value Decomposition Network~\protect\cite{VDN} for RL agent in two client selection tasks. Right: ternary depiction of the 3D optimization space.}
    \label{fig:rl}
    \vspace{-4.5mm}
\end{figure}

\subsection{Client Selection in Federated Learning} 
\looseness=-1
Client selection in FL is critical to optimizing communication costs, computational resources, and overall performance~\cite{oort},
which falls into two main categories:
(1) \textit{heuristic-based selection} that predominantly relies on heuristics rooted in isolated considerations such as data heterogeneity and energy efficiency~\cite{energy_efficient,cho2020client}. 
Specifically, Oort~\cite{oort} employs analytical strategies that comprehensively address the multifaceted nature of device selection. 
However, adapting these heuristics to unseen scenarios often demands lots of domain-specific expertise and extensive tuning.
(2) \textit{learning-based selection} that devises policies for device selection through RL~\cite{reinforcement,tian2024ranking,fan2020autofs} to formulate decisions as Markov decision processes. 
Specifically, AutoFL~\cite{kim2021autofl} leverages a Q-table, Favor~\cite{Favor} adopts Q-learning, and FedMarl~\cite{fedmarl} utilizes multi-agent RL to achieve their objectives.
However, training RL agents is difficult to converge~\cite{dulac2021challenges,fan2021interactive}, especially in the face of noisy and complex real-world environments (as shown in Figure~\ref{fig:rl} left).
Finally, as shown in Figure~\ref{fig:rl} right, all these methods mentioned above usually focus on only one or two optimization metrics and lack a comprehensive global consideration of performance, latency, and energy.

\begin{figure*}[!t]
    \centering
    \includegraphics[width =0.95\linewidth]{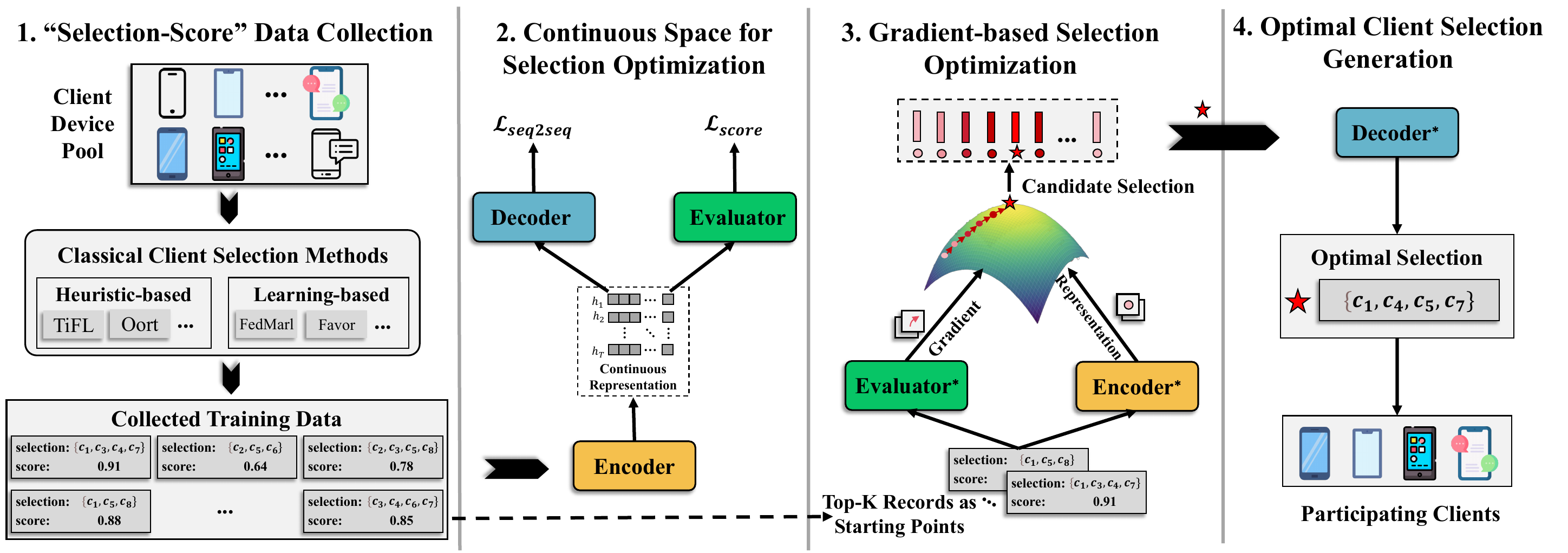}
    \vspace{-1.5mm}
    \caption{Framework overview of \model~: (1) efficiently collecting sufficient, diverse, comprehensive and high-quality training data; (2) preserving the knowledge of classical client selection methods into a global continuous representation space; (3) searching for better representation in the learned space via gradient-based optimization; (4) outputting the optimal device subset via generation.}
    \label{fig:framework}
    \vspace{-4.5mm}
\end{figure*}

\subsection{Generative Models}
\looseness=-1
Recent advancements in LLMs and generative models mark a shift in the machine learning paradigm, attesting to the evolution from task-specific architectures~\cite{ning2022graph,ning2025rethinking,ning2025deep} to highly generalized generic models.
These models, exemplified by the likes of GPT-3~\cite{brown2020language} in the language domain, along with DALL·E~\cite{ramesh2022hierarchical} in the multi-modal domain, have driven the migration to generative models across various domains~\cite{gruver2023large,xiao2023beyond,wang2024reinforcement}.
Generative models have showcased an unparalleled ability to encode multifaceted knowledge representations from expansive data into their parameters~\cite{brown2020language}. 
This encoded knowledge can be adeptly maneuvered to generate contextually coherent and semantically rich outputs, addressing a wide spectrum of downstream tasks~\cite{ning2021lightcake}. 
Despite the wide range of applications in language and vision, we have not seen generative models applied extensively in decision-making tasks like the client selection task in FL.

\section{Problem Formulation}
In this section, we formulate the discrete client selection task in FL as a generative task and conduct gradient-based optimization in a continuous representation space to get the optimal selection. 
Firstly, at the beginning of a training round, given a candidate client device pool set $\mathcal{C} = \{c_1, c_2, \dots \, c_{J}\}$ of $J$ client devices, we use classic client selection algorithms to collect $n$ ``selection-score'' pair as training data for subsequent model training.
We denote the collected records as $\mathcal{R} = \{(\mathbf{s_{i}}, p_{i}) \}_{1}^{n}$, where $\mathbf{s_{i}} = \{s_1, s_2, \dots \, s_{T}\}$ is the client selection record made by a classic algorithm ($\mathbf{s_{i}} \subset \mathcal{C}$, $s_{1 \leq i \leq T} \in \mathcal{C}$ is the selected device IDs, and $T$ is not a fixed number that is determined by the classical algorithm used for this selection), and $p_{i}$ is the corresponding comprehensive score of the FL performance for making the client selection $\mathbf{s_{i}}$.
To characterize $p_{i}$ of a given $\mathbf{s_{i}}$, unlike previous FL approaches~\cite{energy_efficient,zhan2020experience,Favor} that focus on different metrics and typically consider only one or two metrics to guide the selection, in this paper, we simultaneously focus on: 
(1) \textit{good performance} of the global model; 
(2) \textit{low processing latency};
(3) \textit{low energy consumption} to perform comprehensive and multi-dimensional optimization.
Here, the comprehensive metric function $\mathit{M}$ of a client selection $\mathbf{s_{i}}$ is defined as: 
\begin{equation}
    \scalemath{0.85}{
    p_{i} =  \mathit{M}(\mathbf{s_{i}})=p_{perf.} \times\left(\frac{L}{p_{L}}\right)^{\textbf{1}\left(L < p_{L}\right) \times a}
    \times\left(\frac{E}{p_{E}}\right)^{\textbf{1}\left(E < p_{E}\right) \times b}},
    \label{equation:metric}
\end{equation}
\looseness=-1
where $p_{perf.}$ denotes the downstream task performance. 
$L$ and $E$ are developer-specified latency and energy budgets for the devices. 
$p_{L}$ and $p_{E}$ represent the actual total latency and energy consumption, encompassing communication and computation processes, for a given client selection $\mathbf{s_{i}}$.
$\textbf{1}(x)$ is an indicator function that takes the value 1 if $x$ is true and 0 otherwise. 
Therefore, client selections exceeding the desired latency $L$ and energy $E$ will be penalized by developer-specified factors $a$ and $b$~\cite{oort}, both set to 2 in our implementation.
Then, we aim to learn a continuous representation space $\mathbb{S}$ for ``selection-score'' pair data.
Specifically, we learn three modules: 
(1) an encoder $\phi$ which can map a selection $\mathbf{s_{i}}$ to its corresponding continuous representation $E_{\mathbf{s_{i}}}$; 
(2) an evaluator $\omega$ which can evaluate the comprehensive score $p_{i}$ of a selection $\mathbf{s_{i}}$ based on its representation $E_{\mathbf{s_{i}}}$; 
(3) a decoder $\psi$ which can map a continuous representation $E_\mathbf{s_{i}}$ back to its corresponding discrete client selection $\mathbf{s_{i}}$.
We optimize all these modules simultaneously based on the collected records $\mathcal{R}$.
After getting the space $\mathbb{S}$ and the well-learned three modules, we can adopt gradient-based optimization in $\mathbb{S}$ to find the optimal client selection $\mathbf{s^{*}}$, given by:
\begin{equation}
    \begin{aligned}
    \mathbf{s}^*=\psi\left(E_{\mathbf{s^*}}\right)=\psi(\operatorname{argmax}_{E_{\mathbf{s_{i}}} \in \mathbb{S}}\omega(E_{\mathbf{s_{i}}})).
    \end{aligned}
\end{equation}
\vspace{-3mm}
\section{Methodology}
\label{section: framework}
\subsection{Framework Overview}
Figure~\ref{fig:framework} shows the overview of our framework, which consists of four steps: 
(1) \textbf{``selection-score'' data collection}: we collect client selections and their corresponding comprehensive scores as training data for subsequent models;
(2) \textbf{continuous space for selection optimization}: we develop an encoder-evaluator-decoder framework to learn a continuous representation space $\mathbb{S}$ for selection optimization based on the pair data collected in Step 1;
(3) \textbf{gradient-based selection optimization}: we adopt gradient-based optimization in $\mathbb{S}$ to find the optimal client selection representation;
(4) \textbf{optimal client selection generation}: we use the well-trained decoder to generate the optimal client selection on the basis of the optimal representation obtained in Step 3.

\subsection{``Selection-Score'' Data Collection}
To ensure that subsequent models and algorithms perform well, we need to collect sufficient, diverse, comprehensive and high-quality ``selection-score'' pair data $\mathcal{R} = \{(\mathbf{s_{i}}, p_{i}) \}_{1}^{n}$ as training data:
training data is large enough to represent the entire distribution; training data includes high-performance selection cases as well as certain random exploration samples.
Intuitively, we can use random selections and compute their corresponding comprehensive scores as training data. 
However, this strategy is inefficient because the number of random selections is exceptionally large and most of them are low-performance samples.
Therefore, we also must consider the efficiency of our data collection process.

Classical client selection algorithms are capable of outputting well-performing selection samples, and they differ in their methods and the metrics of concern.
With the help of these algorithms, we can automatically and efficiently collect training data that fulfills our requirements in two ways:
(1) \textit{heuristic-base collection:} using heuristic-base selection methods (e.g., Oort~\cite{oort}) to generate records. 
This category of algorithms consists of expert manually-engineered heuristic rules, and different algorithms reflect different experts' considerations, based on which we can produce comprehensive and high-quality ``selection-score'' records reflecting various expert intelligence.
(2) \textit{learning-based collection:} treating learning-based selection methods (e.g., FedMarl~\cite{fedmarl}) as exploration-based training data collectors.
Such methods utilize RL to train continuously self-optimizing agents which progressively explore device selections and find optimal ones.
This stochastic and self-optimizing learning process can be viewed as an exploratory and automated data collection tool that leverages machine intelligence to help collect diverse yet quality records.
For the collected selection samples, the comprehensive metric function $\mathit{M}$ allows us to obtain their corresponding scores easily, thus finally finishing the pair data collection.

\subsection{Continuous Space for Selection Optimization}
After collecting ``selection-score'' pair data $\mathcal{R}$ that reflects the broad and abundant experience of classical client device selection algorithms,
we train an encoder-evaluator-decoder framework based on $\mathcal{R}$ to embed these records into a continuous representation space $\mathbb{S}$ for follow-up gradient-based selection optimization.
Each point in $\mathbb{S}$ is associated with a client selection and its corresponding comprehensive score, denoted below by $\mathbf{s}$ and $p$, respectively.

\paragraph{Data Augmentation.}
Sequence-to-sequence (seq2seq) architecture~\cite{sutskever2014sequence} is an important machine learning method: it uses an encoder to capture the context of the input sequence and sends it to a decoder, which then produces the final output sequence.
Its flexibility and powerful capabilities have led to a wide range of applications where the data can be naturally organized as a sequence~\cite{raffel2020exploring}.
The encoder and decoder together in our framework essentially belong to the seq2seq paradigm as well, with the particularity that the inputs and outputs (\textit{i.e.}, client selections) are a special kind of sequences---the sets.
Each client selection is a subset of the device pool that contains an unordered sequence of device IDs.
To model the order-independent properties of sets in the seq2seq component, we propose to use data augmentation to increase the diversity of sequences corresponding to the same set.
Specifically, given a selection and its corresponding score from $\mathcal{R}$, we randomly shuffle the device IDs contained in the selection to obtain a new order, then we pair the new shuffled device IDs sequence with the original score and add them to the training data.

\paragraph{Encoder $\phi$.}
The encoder aims to map any given selection $\mathbf{s} = \{s_1, s_2, \dots \, s_{T}\}$ to continuous representation $E_{\mathbf{s}} = \phi(\mathbf{s})$ that is considered as the input context.
Here, we adopt a single layer long short-term memory (LSTM)~\cite{hochreiter1997long} as encoder $\phi$. 
Specifically, we input the device IDs from $\mathbf{s}$ sequentially into the encoder and collect their corresponding output embeddings to form $E_{\mathbf{s}} = [h_1, h_2, \dots, h_{T}] \in \mathbb{R}^{T \times d}$ ($T$ is the length of the selection $\mathbf{s}$, and $d$ is the embedding dimension).

\paragraph{Decoder $\psi$.}
The decoder aims to generate the device IDs of a client selection $\mathbf{s}$ based on $E_{\mathbf{s}}$, denoted by $\mathbf{s} = \psi(E_{\mathbf{s}})$.
Similar to the encoder, we employ a single-layer LSTM to implement the decoder and train it in an autoregressive way~\cite{sutskever2014sequence,brown2020language}. 
Firstly, the initial input of the decoder is the last device ID embedding in $E_{\mathbf{s}}$ (\textit{i.e.}, $h_T$). 
Then, at step $i$ of the decoding process, we can get the current decoder hidden state $\hat{h}_i$ from the LSTM, and utilize dot product attention~\cite{luong-etal-2015-effective} to aggregate the input context $E_{\mathbf{s}}$ from encoder, and finally obtain an enhanced input embedding $\tilde{h}_i$ for this step, defined as:
\begin{equation}
\begin{aligned}
\scalemath{0.95}{
\tilde{h}_i=\sum_{h_j \in E_{\mathbf{s}}} a_{i j} h_j \text {, where } a_{i j}=\frac{\exp \left(\hat{h}_i \cdot h_j\right)}{\sum_{h_k \in E_{\mathbf{s}}} \exp \left(\hat{h}_i \cdot h_k\right)},}
\end{aligned}
\end{equation}
where $a_{i j}$ is the attention weight between $\hat{h}_i$ and $h_j$. 
Later, we concatenate $\hat{h}_i$ and $\tilde{h}_i$ together and fed them into a fully connected layer followed by a softmax layer to produce the predictive distribution of step $i$, formulated as:
\begin{equation}
\begin{aligned}
    P_\psi\left(s_i \mid \mathbf{s}_{<i}, E_{\mathbf{s}} \right)=\frac{\exp \left(W_{s_i} \left[\hat{h}_i ; \tilde{h}_i\right]\right)}{\sum_{c \in \mathcal{C}} \exp \left(W_{c} \left[\hat{h}_i; \tilde{h}_i \right]\right)},
\end{aligned}
\end{equation}
where $s_i \in \mathbf{s}$ is the $i$-th ID in $\mathbf{s}$, $\mathbf{s}_{<i}$ represents the prediction of the previous or initial step, $\mathcal{C}$ is the candidate  client device pool set (\textit{i.e.}, the token set in seq2seq models), and $W$ stand for the parameter of the fully connected layer.
By multiplying the probability in each step, we can derive the distribution of the whole client selection $\mathbf{s}$ as follows:
\begin{equation}
\begin{aligned}
P_\psi(\mathbf{s} \mid E_{\mathbf{s}})=\prod_{t=1}^T P_\psi\left(s_i \mid \mathbf{s}_{<i}, E_{\mathbf{s}} \right).
\end{aligned}
\end{equation}

In order to make the generated sequence similar to the true sequence, we minimize the negative log-likelihood of the distribution, which is defined as:
\begin{equation}
\begin{aligned}
\scalemath{0.90}{
\mathcal{L}_{seq2seq}=-\log P_\psi(\mathbf{s} \mid E_{\mathbf{s}})=-\sum_{t=1}^T \log P_\psi\left(s_i \mid \mathbf{s}_{<i}, E_{\mathbf{s}} \right).}
\end{aligned}
\label{equation:loss_seq2seq}
\end{equation}

\paragraph{Evaluator $\omega$.}
The evaluator aims to evaluate the corresponding comprehensive score $p$ of a selection $\mathbf{s}$ based on its representation $E_{\mathbf{s}}$.
Specifically, we first conduct mean value operation on device ID embeddings $h_{{(.)}}$ in $E_{\mathbf{s}}$ to aggregate the information and obtain the integrated selection embedding $\bar{E}_{\mathbf{s}} \in \mathbb{R}^{d}$.
We then fed $\bar{E}_{\mathbf{s}}$ into the evaluator $\omega$ (a feedforward neural network) to estimate the score, given as $\hat{p} = \omega(\bar{E}_{\mathbf{s}})$.
To minimize the difference between the estimated score $\hat{p}$ and the collected real one $p$, we leverage the Mean Squared Error (MSE), defined by:
\begin{equation}
\begin{aligned}
\mathcal{L}_{score}=\operatorname{MSE}(p, \hat{p})=(p-\hat{p})^2.
\end{aligned}
\label{equation:loss_score}
\end{equation}

\paragraph{Joint Training Loss $\mathcal{L}$.}
We optimize the encoder $\phi$, decoder $\psi$ and evaluator $\omega$ simultaneously by integrating Equation~\ref{equation:loss_seq2seq} (\textit{i.e.}, the seq2seq loss) and Equation~\ref{equation:loss_score} (\textit{i.e.}, the score estimation loss) to construct the joint training loss $\mathcal{L}$:
\begin{equation}
\begin{aligned}
\mathcal{L}=\alpha \mathcal{L}_{seq2seq}+(1-\alpha) \mathcal{L}_{score},
\end{aligned}
\end{equation}
where $\alpha$ is the trade-off hyperparameter to control the contribution of $\mathcal{L}_{seq2seq}$ and $\mathcal{L}_{score}$ during the training process.

\subsection{Gradient-based Selection Optimization}
After obtaining the trained encoder, evaluator, and decoder, we can adapt the gradient-based optimization method in $\mathbb{S}$ to find the optimal client selection.
Good initialization is crucial for the gradient-based optimization approaches~\cite{glorot2010understanding}, so we first select top-$K$ client selections ranked by their comprehensive score $p$ and use the encoder to embed these selections into a continuous representation which will later be used as starting points for subsequent optimization.
Assuming that one starting point representation is $E_{\mathbf{s}}$, in order to obtain a selection that possesses a better comprehensive score, we optimize from $E_{\mathbf{s}}$ towards the gradient direction induced by the evaluator $\omega$:
\begin{equation}
\begin{aligned}
E_{\mathbf{s}}^{+}= E_{\mathbf{s}} + \eta \frac{\partial \omega(E_{\mathbf{s}})}{\partial E_{\mathbf{s}}},
\end{aligned}
\label{equation:gradient}
\end{equation}
where $E_{\mathbf{s}}^{+}$ is the optimized selection representation, $\eta$ is the step size.
The comprehensive score of $E_{\mathbf{s}}^{+}$ is supposed to be better than $E_{\mathbf{s}}$ due to $\omega(E_{\mathbf{s}}^{+}) \geq \omega(E_{\mathbf{s}})$.
For all the $K$ starting points, we perform the above optimization several times to get the set of candidate selection representation $\{\tilde{E}_{\mathbf{s}_{i}}\}_{1}^{K}$.
Finally, we choose the optimal selection representation $E_{\mathbf{s^{*}}}$ in $\{\tilde{E}_{\mathbf{s}_{i}}\}_{1}^{K}$ through the estimated candidates' comprehensive score, \textit{i.e.}, $E_{\mathbf{s^{*}}} = \operatorname{argmax}_{\tilde{E}_{\mathbf{s_{i}}}} \{\omega(\tilde{E}_{\mathbf{s_{i}}})\}_{1}^{K}$.

\subsection{Optimal Client Selection Generation}
To identify the optimal client selection $s^{*}$, we fed $E_{\mathbf{s^{*}}}$ into the well-trained decoder $\psi$, this process can be denoted by: $s^{*} = \psi(E_{\mathbf{s^{*}}})$.
Specifically, we adopt the beam search~\cite{freitag-al-onaizan-2017-beam} to generate the client selection, just like text generation in natural language processing~\cite{sutskever2014sequence}.
Instead of setting the length of the selection in advance, we use the decoder to iteratively generate device IDs until it encounters the stop token \textless EOS\textgreater, which makes our selection process more adaptive.

\section{Experiments}
\subsection{Experimental Setup}
\begin{table*}[!ht]
    \vspace{-2mm}
  \small
  \centering
  \begin{threeparttable}
    \begin{tabular}{c|cc|cc|cc|cc|c|c}
    \toprule
     \multirow{4}{*}{\begin{minipage}{1cm}
Dataset \\ \&Model
\end{minipage} }  &   \multicolumn{8}{c|}{CV Tasks}    &     \multicolumn{2}{c}{NLP Tasks} \\ \cline{2-11}

& \multicolumn{2}{c|}{MNIST} & \multicolumn{2}{c|}{CIFAR10} & \multicolumn{2}{c|}{CINIC10} & \multicolumn{2}{c|}{TinyImageNet} & \multicolumn{1}{c|}{Shakespeare} & \multicolumn{1}{c}{Wikitext}\\ 
      & \multicolumn{2}{c|}{LeNet5 $\uparrow$} &  \multicolumn{2}{c|}{Resnet18 $\uparrow$} & \multicolumn{2}{c|}{VGG16 $\uparrow$}   & \multicolumn{2}{c|}{ShuffleNet $\uparrow$} & \multicolumn{1}{c|}{LSTM $\uparrow$}  & \multicolumn{1}{c}{GPT-2\tnote{1} $\downarrow$} \\ \cline{2-11}
      & IID & Non-IID\tnote{2} & IID & Non-IID\tnote{2} & IID & Non-IID\tnote{2}& IID & Non-IID\tnote{2}& Non-IID\tnote{2} & Non-IID\tnote{2}\\
    \hline  
    FedAvg &    98.84   &    36.70   &   84.78    &   42.02    & 64.78    & 37.40 &    34.62   &  23.82    &  39.18   &   31.13      \\
     FedProx & \underline{$99_{(35)}$}\tnote{3}  &   67.67  &   85.11    &  52.40    &   64.51 & 39.59 &   35.38  &  24.72    &   40.68    &   30.07   \\
    % \hline
    AFL & 98.61  &   89.60  &   83.33   &  51.81   & 65.86   &  41.03 & 36.77  & 25.90    &   42.67   &  27.42       \\
    TiFL &  98.91 &   91.21  &   84.96    &  56.39    &  \underline{67.32}  & 42.16 & 38.82  &  25.85   &   44.12    &     25.61     \\
    Oort & $99_{(49)}$\tnote{3}  &    \underline{92.47}   &   85.69   &  53.51     &  67.07  & 43.57 &  \underline{40.57} &      \underline{26.62}  &    \underline{45.42}  &    \underline{19.12}    \\ 
    % \hline
     Favor & 98.87  &  85.48   &   85.98    &    51.34  & 65.47   & 39.26 &  37.62 &   25.34  &   43.11   &   28.45      \\
    FedMarl &  98.82     &   90.13   &   \underline{86.37}  &   \underline{56.73}   &  66.74    &  \underline{44.16}  &   39.48    &   26.31  &   44.72    &   21.84     \\
    \hline
    \model~ &  $\textbf{99}_{(28)}$\tnote{3}   & \textbf{95.83}     &  \textbf{88.42}    &    \textbf{61.62}   &   \textbf{69.09}   & \textbf{46.27} &  \textbf{43.35} &   \textbf{ 29.51}  &    \textbf{47.65}   &  \textbf{12.58}       \\
    \bottomrule    
    \end{tabular}%
    \vspace{-1.5mm}
  \caption{Performance of different selection methods. \textbf{Bold} indicates the best one and \underline{underline} indicates the runner-up.}
  \label{tab:acc}%
    \begin{tablenotes}
    % \footnotesize 
    % \scriptsize
    \item[1] We use perplexity (PPL) to evaluate GPT-2, which reflects the text generation capability of model.
    \item[2] Set $\beta =0.01$ for MNIST, $\beta =0.1$ for others to emulate Non-IID. Shakespeare and Wikitext are naturally Non-IID.
    \item[3] Early exit when target accuracy ($99\%$) is achieved. Numbers in parentheses indicate the exited round.
    \end{tablenotes}
  \end{threeparttable}
    \vspace{-2mm}
\end{table*}%

\paragraph{Infrastructure.} 
To evaluate the performance and effectiveness of \model~, we first build a simulator with the server/client architecture based on PyTorch~\cite{pytorch}, where distinct processes emulate the central server and participating devices.
To emulate data heterogeneity, we allocate training data with different distributions across devices. 
To mimic system heterogeneity, we establish a FL system comprising six device types with diverse hardware configurations, including Google Pixel 6, OnePlus 10 pro, Redmi Note 10, Realme Q3s, NVIDIA Jetson Nano, and NVIDIA Jetson TX2.
We utilize Monsoon Power Monitor~\cite{monsoon} to track latency and energy consumption during training. 
Furthermore, we integrate end-user interaction traces from LiveLab~\cite{livelab} to emulate concurrent applications that impact the training capability at runtime.

\paragraph{Baselines.} \model~ is compared with three types of client selection methods, 
including \textbf{\textit{random-based}} (FedAvg~\cite{fedavg} and FedProx~\cite{fedprox}), 
\textbf{\textit{heuristic-based}} (AFL~\cite{afl},  TiFL~\cite{tifl} and Oort~\cite{oort}),
and \textbf{\textit{learning-based}} (Favor~\cite{Favor} and FedMarl~\cite{fedmarl}).

\paragraph{Datasets and Models.} We use typical models and datasets in Computer Vision (CV) and Natural Language Processing (NLP) domains for evaluation, 
including LeNet5~\cite{lenet5} on MNIST~\cite{mnist}, 
ResNet18~\cite{resnet} on CIFAR10~\cite{cifar10}, 
VGG16~\cite{vgg16} on CINIC10~\cite{cinic}, 
ShuffleNet~\cite{shufflenet} on TinyImageNet~\cite{imagenet} for image classification, LSTM~\cite{hochreiter1997long} on Shakespeare~\cite{Shakespeare}, 
GPT-2~\cite{gpt2} on Wikitext~\cite{wikitext} for text generation. 
For four CV datasets, Dirichlet distribution $p_k \sim Dir_N(\beta)$ is utilized to simulate the Non-IID data distribution on different devices.

\paragraph{Hyperparameter Settings and Reproducibility.}
We run two heuristic-based methods---Oort and Favor, and one heuristic-based method---FedMarl 100 times each, totaling 300 runs, to collect ``selection-score'' pair data for subsequent model training. 
For data augmentation, each selection is randomly shuffled 25 times to model set order-independence. 
We adopt a single-layer LSTM as the Encoder and Decoder backbones, and two-layer feed-forward networks for the Evaluator. 
The hidden state sizes are 64 (Encoder), 64 (Decoder), and 200 (Evaluator), with a $32$-dimensional embedding for each device ID token. 
For \model~ training, we set batch size$=1024$, learning rate$=0.001$, $\alpha=0.8$, and utilize the top-$25$ device selections as starting points for gradient optimization. 
In the FL setting, $J=100$ devices are available, with $T=10$ ($T=20$ for FedMarl) selected per round over $r=10$ rounds for GPT-2 ($r=50$ for others). 
Clients perform local $ep = 5$ epoch training each round.
The code is available at \url{https://github.com/zhiyuan-ning/GenerativeFL}.

\subsection{Overall Performance}
\paragraph{Model Performance.} Table~\ref{tab:acc} shows the final test performance of all client selection methods.
We can observe that \model~ significantly outperforms other baselines across all domains and tasks. 
Specifically, for IID setting, \model~ improves the test accuracy $5.56\%$ over FedAvg and $2.20\%$ over SOTA value on average. 
Further, \model~ is particularly more effective in the more challenging Non-IID setting, outperforming the baselines by considerable margins: improving the test accuracy $20.35\%$ over FedAvg and $3.16\%$ over SOTA on average, and reducing $18.55$ perplexity (PPL, lower values correspond to stronger text generation capability) over FedAvg and $6.54$ over SOTA on the GPT-2 model.
Overall, this experiment demonstrates the practical and robust ability of \model~ to scale to complex workloads and application scenarios.
This may be because \model~ converts extensive discrete decision knowledge into a continuous representation space in a data-driven manner and identifies a superior client selection in that space based on gradient-based optimization.

\begin{figure}[t]
    \centering
    \vspace{-2mm}
    \includegraphics[width=0.48\linewidth]{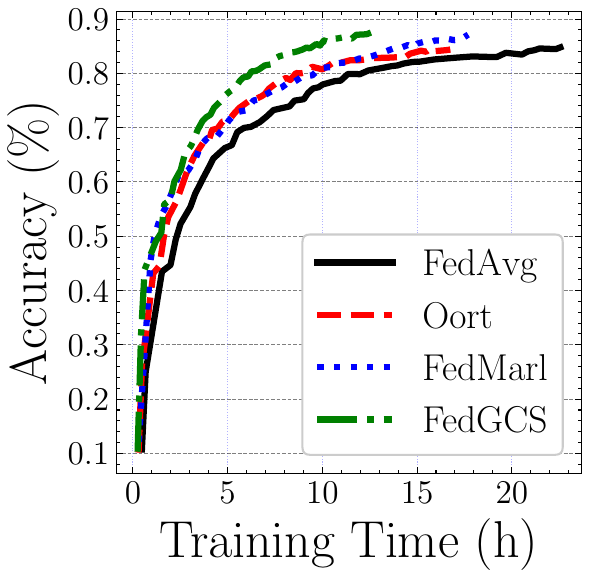}
    \includegraphics[width=0.48\linewidth]{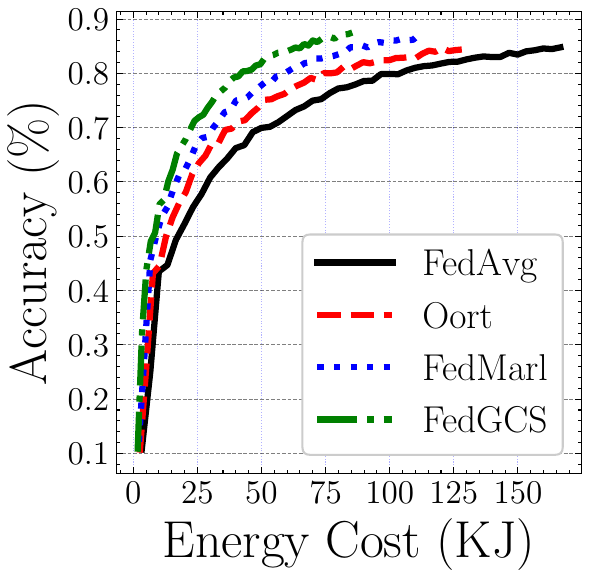}
    \vspace{-2mm}
    \caption{Efficiency comparison of \model~ with three SOTA baselines to train ResNet18 on CIFAR10 (IID). Left: ToA. Right: EoA.
    }
    \label{fig:system}
    \vspace{-4mm}
\end{figure}

\paragraph{Latency and Energy Consumption.}
Then, we evaluate the system efficiency of \model~ from two perspectives: Time to Accuracy (ToA) and Energy to Accuracy (EoA). 
Figure~\ref{fig:system} left shows \model~ effectively speeds up the training process with faster convergence and achieves superior accuracy. 
In addition, \model~ also achieves significant energy savings shown in Figure~\ref{fig:system} right. 
This can be attributed to \model~ thoughtfully integrating  considerations of both latency and energy consumption during training. 
In contrast, Oort emphasizes solely training duration as the system effectiveness metric, and FedMarl takes into account the energy costs associated with communication, but it overlooks the intrinsic energy overheads of the training process itself, a critical oversight, especially for devices operating on battery power.

\begin{table}[!t]
    \centering
    \small
    \begin{tabular}{l|ccc}
    \toprule
        Schemes & Acc. (\%)$\uparrow$ & Latency(min)$\downarrow$ & Energy(KJ)$\downarrow$  \\
    \midrule
        $\text{\model~}^{-c}$ &86.23 & 18.2 & 2.12\\
        $\text{\model~}^{-a}$ & 85.37 & 24.82 & 2.53 \\
        \textbf{\model~} &  \textbf{88.42} & \textbf{15.34} & \textbf{1.70} \\
    \bottomrule
    \end{tabular}
    \vspace{-1mm}
    \caption{The effect of data collection ($\text{\model~}^{-c}$) and augmentation ($\text{\model~}^{-a}$) on test accuracy, average latency and energy cost per round when training the ResNet18 model on CIFAR10 (IID).}
    \vspace{-5mm}
    \label{tab:ablation}
\end{table}

\subsection{Framework Analysis}
\paragraph{Impact of Data Collection and Augmentation.} 
To explore the impact of the different modules of \model~, we develop two model variants: 
(1) $\text{\model~}^{-c}$, we randomly collect pair data without relying on classical client selection methods. 
(2) $\text{\model~}^{-a}$, we disable the data augmentation process of \model~. 
Table~\ref{tab:ablation} shows the comparison results, we can find that the performance of \model~ is much better than $\text{\model~}^{-c}$. 
This suggests that the quality of the collected data is critical to constructing a continuous representation space, and that a better space will in turn help to identify better client selection.
Moreover, we observe that \model~ is superior to $\text{\model~}^{-a}$. 
The key driver is that data augmentation can enhance data diversity and model the order-independent properties of client selection, thereby enabling more robust and effective learning for \model~. 
Consequently, these results confirm that data collection and augmentation are essential for sustaining the performance of \model~.

\begin{figure}[!ht]
    \centering
    \includegraphics[width =1.0\linewidth]{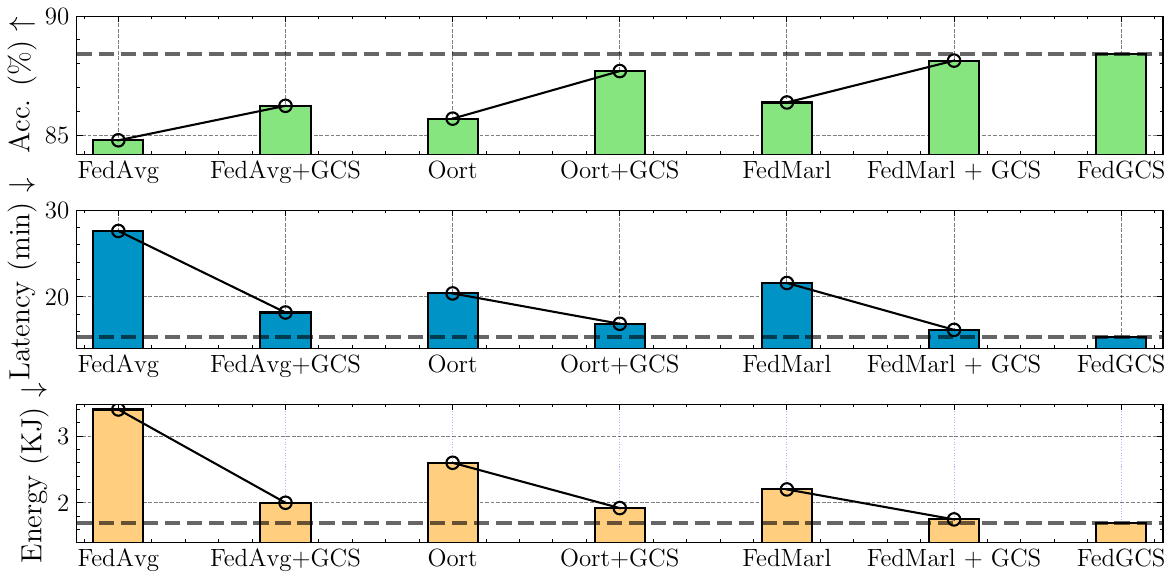}
    \vspace{-5mm}
    \caption{Generalization ability of \model~.}
    \label{fig:gen}
    \vspace{-2mm}
\end{figure}

\paragraph{Study of Generalization Ability of GCS.} 
Figure~\ref{fig:gen} outlines when training the ResNet18 model on CIFAR10 (IID), the generalization ability of \model~ in combination with different client selection methods, and the effectiveness of using multiple selection methods for data collection. 
The adoption of a single selection method as data collector can significantly improve results over the corresponding selection method, highlighting the merits of continuous space paradigms in enhancing generalization ability.
Furthermore, the use of diverse, multiple selection methods as data collectors (\textit{i.e.}, the \model~ in Figure~\ref{fig:gen} which uses Oort, Favor and FedMarl as data collectors simultaneously) outperforms other single collector strategies, emphasizing the value of diversity in data collectors for comprehensive space construction.

\begin{figure}[!ht]
    \centering
    \includegraphics[width =1.0\linewidth]{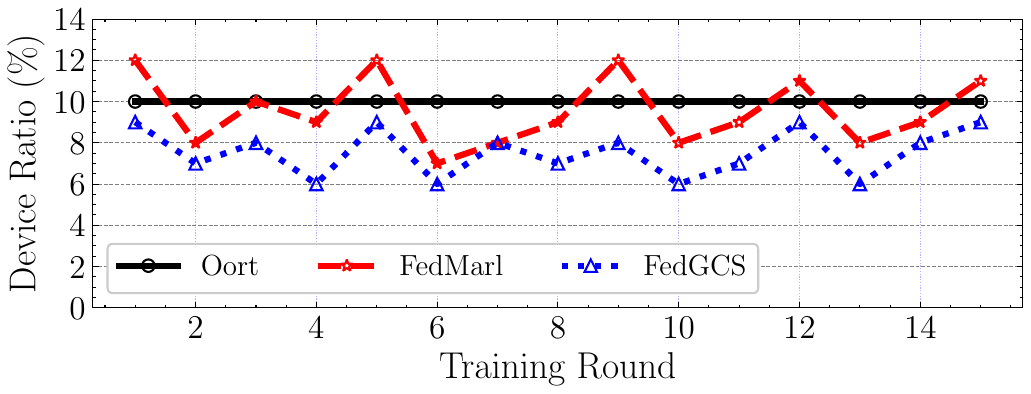}
    \vspace{-6mm}
    \caption{Comparison of the selected devices ratio of \model~, FedMarl and Oort.}
    \label{fig:select_1}
    \vspace{-4mm}
\end{figure}

\paragraph{Study of the Selection Strategy Made by \model~.} 
We analyze the selection strategy of \model~ by comparing its selected device ratio with the best fixed client selection method Oort and the dynamic selection approach FedMarl.
Figure~\ref{fig:select_1} shows the results, where we can observe that the ratio of the client selection made by \model~ is dynamically changing each training round, indicating that decision making via generative is more adaptive when compared to the fixed client selection methods.
In addition, the ratio of \model~ is the smallest, suggesting that \model~ can select the most reasonable number of devices even when compared to the dynamic selection approaches.
Overall, such results demonstrate that \model~ can achieve the best performance with the smallest device selection ratio, indicating that the selection strategy of \model~ is superior compared to other methods.

\begin{figure}[!ht]
    \centering
    \subfigure[Trade Off]{\includegraphics[width=0.46\linewidth]{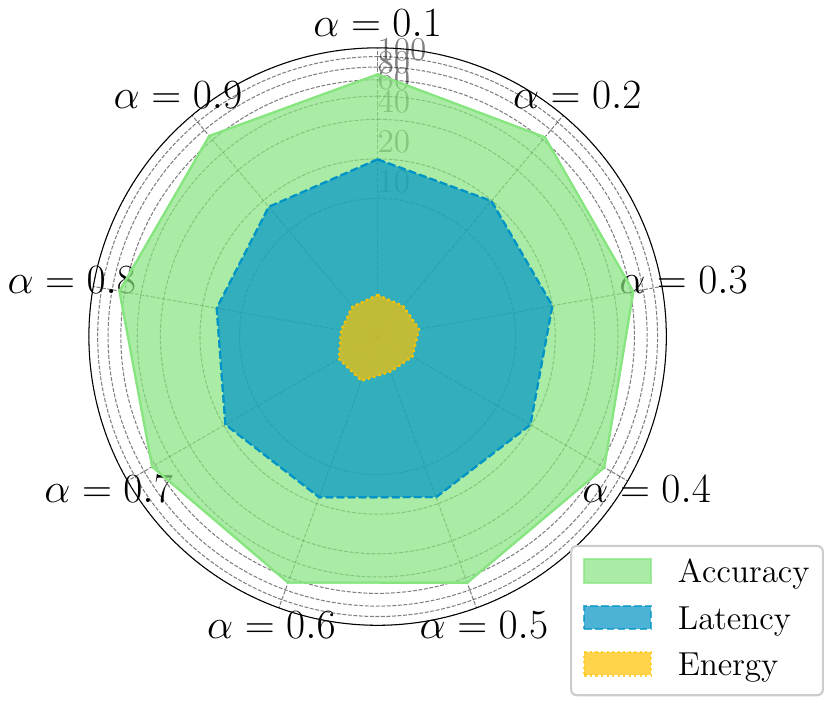}}
    \hfill
    \subfigure[Top-K]{\includegraphics[width=0.46\linewidth]{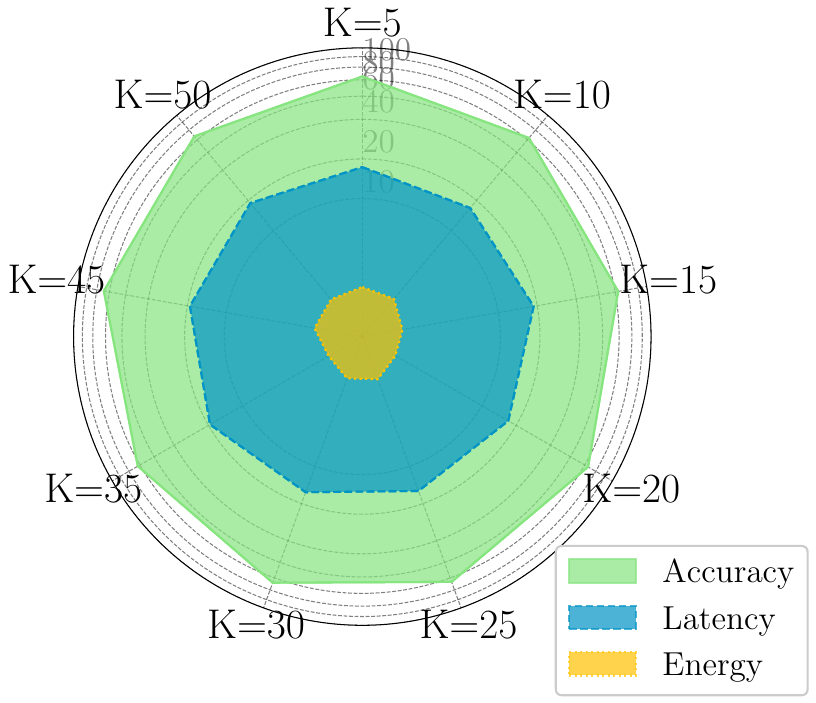}}
    \vspace{-4mm}
    \caption{Hyperparameter sensitivity of \model~.}
    \label{fig:sensitivity}
    \vspace{-3mm}
\end{figure}

\paragraph{Study of the Hyperparameter Sensitivity of \model~.} 
There are two major hyperparameters, training trade-off $\alpha$ and the top-$K$ records as the starting points of gradient-based optimization. 
A higher $\alpha$ will make the model more concentrated on the loss $\mathcal{L}_{seq2seq}$, and a higher $K$ makes the model search more milder. 
We set $\alpha$ from 0.1 to 0.9, and $K$ from 5 to 50, then train the \model~ on ResNet18 over CIFAR10 (IID). 
The model performance is reported in Figure~\ref{fig:sensitivity}. 
Overall, we observe that model performance and system efficiency vary slightly for different K, but for better results K needs to be greater than 20.
Further, setting $\alpha$ as 0.8 will slightly bring a higher model performance. 
These findings provide insight into how $\alpha$ and $K$ impact the model performance and system efficiency and how to choose the optimal hyperparameters. 

\begin{figure}[!ht]
    \centering
    \includegraphics[width =1.0\linewidth]{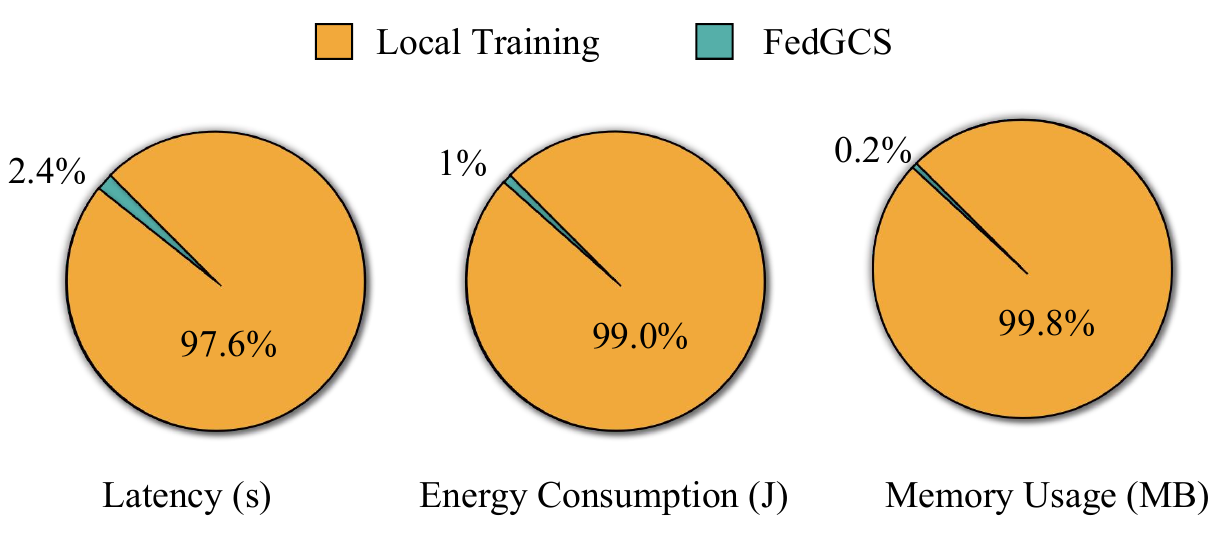}
    \vspace{-6mm}
    \caption{Overhead analysis of the \model~ framework shows low memory usage and operational effectiveness.}
    \label{fig:overhead}
    \vspace{-3mm}
\end{figure}

\subsection{Overhead Analysis}
As shown in Figure~\ref{fig:overhead}, we delineate the overhead analysis of the \model~ framework and employ the Monsoon Power Monitor for the empirical measurement of the local training runtime overhead, specifically focusing on the ResNet18-CIFAR10 model augmented with 500 samples on an NVIDIA Jetson Nano platform. 
The empirical data reveals an average local training duration per round of 28.24 minutes, accompanied by an average energy expenditure of 370.5 Joules. Conversely, the deployment of the \model~ framework on a server manifests markedly different metrics: a mere 0.68 minutes of processing time and an average energy consumption of 3.9 Joules. This stark contrast underscores the negligible runtime overhead of the \model~ when operating in a server environment.
Further, the memory requirements of \model~ are remarkably low, with a total memory footprint under 0.1 MB. This is negligible when compared to the substantial memory size of the ResNet18 model, which is approximately 42.70MB, and the expansive 40GB DRAM capacity available on cloud servers equipped with NVIDIA A40 GPUs. Consequently, we assert that the runtime overhead of the \model~ framework is not a significant factor in affecting the overall efficiency and progress of the training process.

\vspace{-3mm}
\section{Conclusion}
\model~ marks a significant advancement in FL by introducing a generative framework for client selection, effectively addressing statistical and system heterogeneity and high energy consumption challenges. 
This framework transcends traditional methods by encapsulating decision-making processes within a continuous representation space, enabling precise and efficient identification of optimal client subsets. 
Our novel approach, which includes data collection, model training, gradient-based optimization, and generation of client selection, not only outperforms existing methods in terms of model performance, latency, and energy efficiency but also simplifies the process, reducing reliance on domain expertise and manual intervention. 
The empirical validation of \model~ underscores its superiority and potential to revolutionize client selection strategies in FL.

\section*{Acknowledgments}
This research is supported by the 
the Strategic Priority Research Program of the Chinese Academy of Sciences XDB38030300, 
the Natural Science Foundation of China under Grant No. 61836013, 
the Postdoctoral Fellowship Program of CPSF (No.GZC20232736),
the China Postdoctoral Science Foundation Funded Project (No.2023M743565),
the Science and Technology Development Fund (FDCT),
Macau SAR (file no. 0123/2023/RIA2, 001/2024/SKL),
and the Start-up Research Grant of University of Macau (File no. SRG2021-00017-IOTSC).

% \clearpage
\appendix
\section*{Appendix}

\section{The Detailed Process of Data Collection}
In our implementation, we use one heuristic-base selection method: Oort~\cite{oort}, and two learning-based selection methods: Favor~\cite{Favor} and FedMarl~\cite{fedmarl} as our training data collectors to obtain appropriate selections.
They differ in the metrics they focus on and the specific manner of their realization.

For Oort, it optimizes device selection by defining the utility of client $i$ as the product of the statistical utility and the global system utility, which integrates training loss and latency to maximize per-unit-time efficiency.
\begin{equation}
\operatorname{Util}(i)=\underbrace{\left|B_i\right| \sqrt{\frac{1}{\left|B_i\right|} \sum_{k \in B_i} \operatorname{Loss}(k)^2}}_{\text {Statistical utility } U(i)} \times \underbrace{\left(\frac{T}{t_i}\right)^{\textbf{1}\left(T<t_i\right) \times \alpha}}_{\text {Global sys utility }}
\end{equation}

For Favor, it employs a double DQN-trained DRL agent to select local weights based on accuracy, countering Non-IID challenges and boosting convergence, optimizing rewards represented by:
\begin{equation}
    R=\sum_{t=1}^T \gamma^{t-1}\left(\Xi^{\left(\omega_t-\Omega\right)}-1\right)
\end{equation}

For FedMarl, it employs multi-agent reinforcement learning for device selection in federated learning, optimizing accuracy, latency, and communication cost, with rewards per round defined as: 
\begin{equation}
r_t=w_1[U(Acc(t))-U(Acc(t-1))]-w_2 H_t-w_3 B_t
\end{equation}

We ran each of the three training data collectors 100 times, thus collecting 300 client selections.
For those diverse, comprehensive, and high-quality selections, we devise the following formula that combines performance, latency, and energy consumption to obtain the corresponding comprehensive scores $p$ for each collected selection, and thus the final ``selection-score'' training data:

\begin{equation}
    \begin{aligned}
        p &= p_{Perf} \times  (\frac{L}{p_{L}})^{\textbf{1}(L < p_{L})\times \alpha} \times (\frac{E}{p_{E}})^{\textbf{1}(E < p_{E})\times \beta} \\
        &\text{where:} \\
        p_{L} &= \max_{0 \leq i \leq T}\{[L_{comm,i}+L_{comp,i}*l_{ep}]\}\\
        p_{E} &= \sum_{T}\{[E_{comm,i}+E_{comp,i}*l_{ep}]\}
        \label{opti_problem}
    \end{aligned}
\end{equation}
where $L$ and $E$ are the developer-preferred duration and energy budget of the devices, respectively. 
\textit{comm} represents the communication process and \textit{comp} represents the computation process.
$l_{ep}$ is the number of local training epochs. 
$\textbf{1}(x)$ is an indicator function that takes the value 1 if $x$ is true and 0 otherwise. 
In this way, the utility of those clients that may be the bottleneck of the desired speed and energy cost of the current round is penalized by a developer-specified factor $\alpha$ and $\beta$, but we do not reward the non-straggler and non-overloader clients because their completions do not affect the effectiveness of the training round~\cite{oort}.
We set both $\alpha$ and $\beta$ to 2 in our implementation.

\section{Experiment Details}

\subsection{Dataset Details}
We use four computer vision datasets and two natural language processing datasets to fully validate the capabilities of our methodology.
Following are their detailed descriptions:
\begin{itemize}[leftmargin=*]
  \item \textbf{MNIST}~\cite{mnist} is a large collection of handwritten digits. It has a training set of $60,000$ examples and a test set of $10,000$ examples. The images are centered in a $28\times28$ image by computing the center of mass of the pixels, and translating the image so as to position this point at the center of the $28\times28$ field.
  \item \textbf{CIFAR10}~\cite{cifar10} is a subset of the Tiny Images dataset and consists of $60,000$ $32\times32$ color images. These images are labeled into one of ten mutually exclusive categories: airplanes, automobiles (but not trucks or pickups), birds, cats, deer, dogs, frogs, horses, ships, and trucks (but not pickups). There are $6000$ images for each category, with 5000 training images and 1000 test images.
  \item \textbf{CINIC10}~\cite{cinic} is a dataset for image classification that has a total of $270,000$ images. It is constructed from two different sources: ImageNet and CIFAR10. It is split into three equal subsets---train, validation, and test---each of which contains $90,000$ images.
  \item \textbf{TinyImageNet}~\cite{imagenet} contains $100,000$ images of $200$ classes ($500$ for each class) downsized to $64\times64$ colored images. Each class has $500$ training images, $50$ validation images, and $50$ test images.
  \item \textbf{Shakespeare}~\cite{Shakespeare} is a dataset built from The Complete Works of William Shakespeare. It is a popular choice for training language models due to its manageable size and the complexity of Shakespeare's language. It provides a good balance between computational efficiency and the ability to generate interesting text.
  \item \textbf{WikiText}~\cite{wikitext} is a language modeling dataset that is a collection of over 100 million tokens extracted from the set of verified Good and Featured articles on Wikipedia. The dataset is available under the Creative Commons Attribution-ShareAlike License.
\end{itemize}

For four computer vision datasets, we generate IID data splits by randomly assigning training examples to each client without replacement. 
For Non-IID splits, we simulate data heterogeneity by sampling label ratios from a Dirichlet distribution $p_k \sim Dir_N(\beta)$ with the symmetric parameter $\beta$~\cite{hsu2019measuring}.
We set $\beta =0.01$ for MNIST, and $\beta =0.1$ for others to emulate Non-IID.
For two natural language processing datasets, Shakespeare and Wikitext are naturally non-IID.
We use LeNet5~\cite{lenet5} on MNIST, ResNet18~\cite{resnet} on CIFAR10, VGG16~\cite{vgg16} on CINIC10, ShuffleNet~\cite{shufflenet} on TinyImageNet, LSTM~\cite{hochreiter1997long} on Shakespeare, GPT-2~\cite{gpt2} on
Wikitext.
For a fair comparison, baselines and \model~ will be compared under the same settings.

\begin{table*}[!ht]
  \small
  \centering
  \begin{threeparttable}
  % \caption{Model performance of different selection approaches. \model~ achieves the best performance across all the datasets.}
    \begin{tabular}{c|cc|cc|cc|cc|c|c}
    \toprule
     \multirow{4}{*}{\begin{minipage}{1cm}
Dataset \\ \&Model
\end{minipage} }  &   \multicolumn{8}{c|}{CV Tasks}    &     \multicolumn{2}{c}{NLP Tasks} \\ \cline{2-11}

& \multicolumn{2}{c|}{MNIST} & \multicolumn{2}{c|}{CIFAR10} & \multicolumn{2}{c|}{CINIC10} & \multicolumn{2}{c|}{TinyImageNet} & \multicolumn{1}{c|}{Shakespeare} & \multicolumn{1}{c}{Wikitext}\\ 
      & \multicolumn{2}{c|}{LeNet5} &  \multicolumn{2}{c|}{Resnet18} & \multicolumn{2}{c|}{VGG16}   & \multicolumn{2}{c|}{ShuffleNet} & \multicolumn{1}{c|}{LSTM}  & \multicolumn{1}{c}{GPT-2} \\ \cline{2-11}
      & IID & Non-IID\tnote{2} & IID & Non-IID\tnote{2} & IID & Non-IID\tnote{2}& IID & Non-IID\tnote{2}& Non-IID\tnote{2} & Non-IID\tnote{2}\\
    \hline  
    FedAvg &    9.84   &  10.21     &   28.24    &   31.67    &  67.89   &  71.23 &    81.43   &  85.12   &  163.32   &   244.12     \\
     FedProx & 9.57  &   9.78  &  27.87   &  30.45    &  64.52  &  67.71 &   78.01  &   82.32  &    154.18  &   234.41   \\
    % \hline
    AFL & 8.92  &  9.01  &  26.21  &  28.91   & 63.45  & 60.91  &  70.96 &  78.43  &   145.76   &    230.65     \\
    TiFL & 8.21  &  8.47   &    25.43   &   26.22  &  50.24  &  53.08&  61.23 &  67.12   &   128.94   &    190.82    \\
    Oort &  \underline{7.84}  &   \underline{7.52}   &  \underline{20.13}   &   \underline{23.01}    &  41.89  &  48.65 &  53.78 &   56.45     &    \underline{108.38}  &    168.23   \\ 
    % \hline
     Favor   &  9.51  &   9.44  &    25.62  &  27.32  & 58.98 &  59.71  &  65.78   &  71.07  &  149.45      &  195.64\\
    FedMarl &   9.16   &   8.82  &  21.60   &   23.42   &   \underline{38.23}   &  \underline{42.54}  &   \underline{51.08}    &   \underline{54.09}  &   119.83    &    \underline{150.76}   \\
    \hline
    \model~ &  \textbf{6.13}   &  \textbf{6.48}    &  \textbf{ 15.34}   &   \textbf{18.76}   &  \textbf{32.28}    & \textbf{40.19} &  \textbf{43.98} &  \textbf{ 44.42}  &   \textbf{94.59}   &    \textbf{138.71}     \\
    \bottomrule    
    \end{tabular}%
  \caption{The average latency (min) of each training round for different selection approaches. \model~ achieves the best performance across all the datasets.}
  \label{tab:latency}%
    \begin{tablenotes}
    % \footnotesize 
    \scriptsize
    \item[1] Set $\beta =0.01$ for MNIST, $\beta =0.1$ for others to emulate Non-IID. Shakespeare and Wikitext are naturally non-IID.
    \end{tablenotes}
  \end{threeparttable}
\end{table*}%

\begin{figure*}[!ht]
    \centering
    \subfigure[MNIST (IID)]{\includegraphics[width=0.325\linewidth]{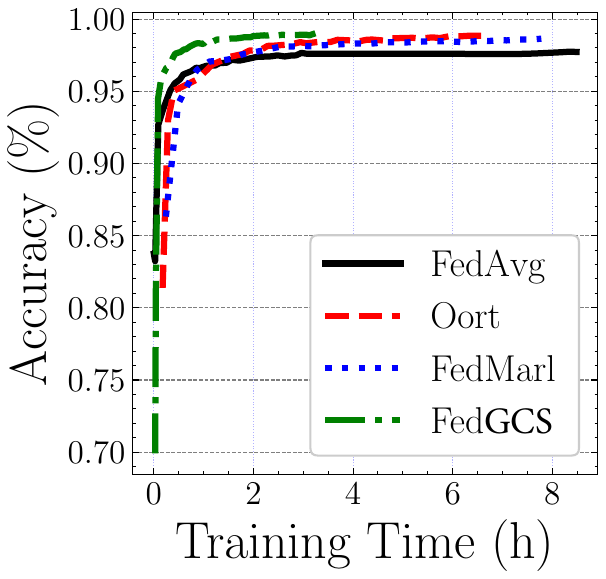}}
    \subfigure[MNIST (NonIID)]{\includegraphics[width=0.315\linewidth]{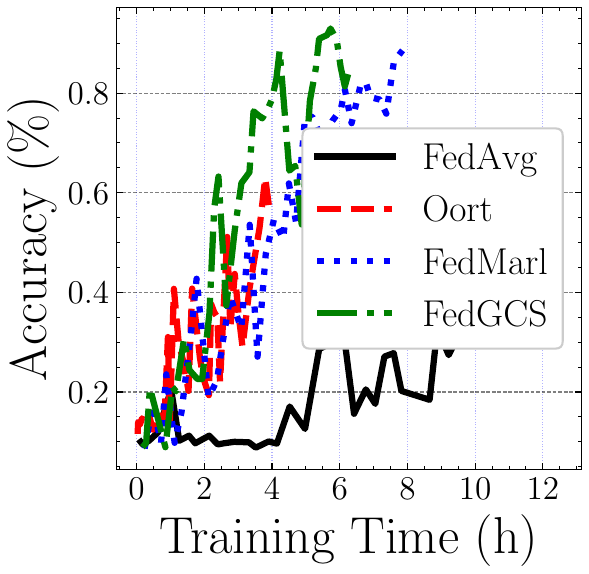}}
    \subfigure[CIFAR10 (IID)]{\includegraphics[width=0.315\linewidth]{Figures/latency.pdf}}
    \subfigure[CIFAR10 (NonIID)]{\includegraphics[width=0.315\linewidth]{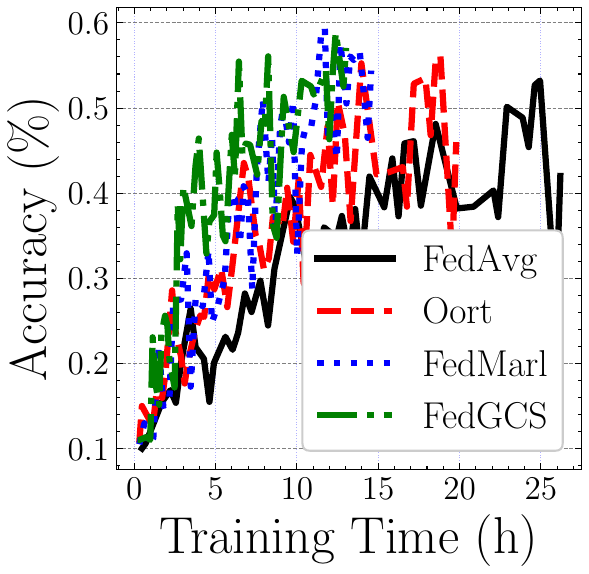}}
    \subfigure[CINIC10 (IID)]{\includegraphics[width=0.315\linewidth]{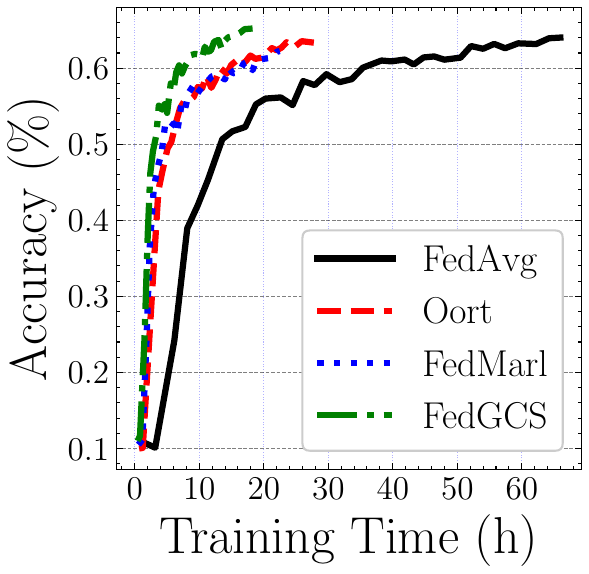}}
    \subfigure[CINIC10 (NonIID)]{\includegraphics[width=0.325\linewidth]{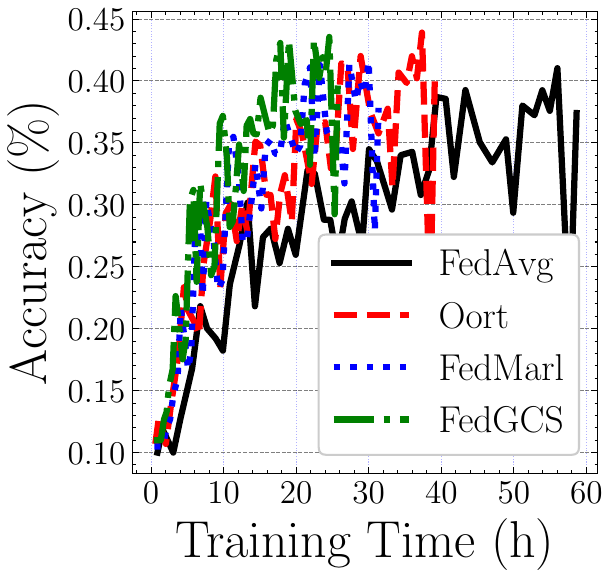}}
    \vspace{-4mm}
    \caption{ToA: time to accuracy. Efficiency comparison of various schemes.  
    }
    \label{fig:system-latency}
\end{figure*}

\subsection{Baseline Details}
We compare \model~ with three types of representative device selection approaches, including 
\begin{itemize}[leftmargin=*]
\item Random Selection:
\begin{itemize}
\item \textbf{FedAvg}~\cite{fedavg}, a vanilla framework for FL without any operation. In each round \( t \), the server randomly selects a subset of available devices \( K_t \) from the total devices \( K \). Each chosen device \( k \) trains the model on its local dataset using the current global model parameters \( w_t \), resulting in updated local parameters \( w_{t+1}^k \).
The server aggregates these updated local parameters using the formula:
\begin{equation}
    w_{t+1} = \frac{1}{K_t} \sum_{k=1}^{K_t} w_{t+1}^k
\end{equation}
This step effectively averages the local updates to form the new global model parameters.

\item \textbf{FedProx}~\cite{fedprox} improves FedAvg by handling heterogeneous data and systems, adding a proximal term to the local training objective to address non-IID data and system differences. In each round \( t \), a subset of devices \( K_t \) is selected to train on their local data with the global model parameters \( w_t \) and a proximal term. The local update \( w_{t+1}^k \) is obtained by minimizing \( L_k(w) + \frac{\mu}{2} \|w - w_t\|^2 \), where \( L_k(w) \) is the local loss, and \( \mu \) adjusts the proximal term's influence. FedProx's main advantage is its ability to stabilize training across diverse data and system environments by ensuring local updates remain close to the global model.
\end{itemize}

\item Heuristic-based Selection:
\begin{itemize}
\item \textbf{AFL}~\cite{afl} uses a device selection method conditioned on the model and client data to enhance efficiency in each round \( t \). This method focuses on devices likely to offer significant model improvements based on data diversity, model uncertainty, or past updates' impact. Post-training, devices submit their updates and informativeness measures (like loss or uncertainty) to the server. The server then combines these updates, potentially weighted by informativeness, to revise the global model to \( w_{t+1} \), the following:
\begin{equation}
   w_{t+1} = \sum_{k=1}^{K_t} \alpha_k w_{t+1}^k
\end{equation}
Here, \( \alpha_k \) denotes the weight of the \( k \)-th device's contribution.

\item \textbf{TiFL}~\cite{tifl} groups devices by response latency to mitigate system heterogeneity, sorting them into tiers by capability, bandwidth, and data quality. Each tier has a Tier Server (TS) where selected devices contribute to local updates. These updates are first aggregated within tiers at the TS using:
\begin{equation}
   w_{t+1}^{TS} = \frac{1}{K_t} \sum_{k=1}^{K_t} w_{t+1}^k
\end{equation}
Then, Tier Servers forward their updates to a central server for global aggregation, potentially weighted by tier attributes:
\begin{equation}
   w_{t+1} = \sum_{i=1}^{T} \beta_i w_{t+1}^{TS_i}
\end{equation}
This balances individual contributions and overall data characteristics.

\item \textbf{Oort}~\cite{oort} optimizes device selection by integrating training loss and latency into a user-defined utility. To enhance efficiency, it's crucial to maximize the per-unit-time statistical utility. The utility of client $i$ is defined as a product of her statistical utility and the global system utility, considering the duration of each training round, as shown in the equation:
\begin{equation}
\operatorname{Util}(i)=\underbrace{\left|B_i\right| \sqrt{\frac{1}{\left|B_i\right|} \sum_{k \in B_i} \operatorname{Loss}(k)^2}}_{\text {Statistical utility } U(i)} \times \underbrace{\left(\frac{T}{t_i}\right)^{\textbf{1}\left(T<t_i\right) \times \alpha}}_{\text {Global sys utility }}
\end{equation}
\end{itemize}

\begin{table*}[!ht]
  \small
  \centering
  \begin{threeparttable}
  % \caption{Model performance of different selection approaches. \model~ achieves the best performance across all the datasets.}
    \begin{tabular}{c|cc|cc|cc|cc|c|c}
    \toprule
     \multirow{4}{*}{\begin{minipage}{1cm}
Dataset \\ \&Model
\end{minipage} }  &   \multicolumn{8}{c|}{CV Tasks}    &     \multicolumn{2}{c}{NLP Tasks} \\ \cline{2-11}

& \multicolumn{2}{c|}{MNIST} & \multicolumn{2}{c|}{CIFAR10} & \multicolumn{2}{c|}{CINIC10} & \multicolumn{2}{c|}{TinyImageNet} & \multicolumn{1}{c|}{Shakespeare} & \multicolumn{1}{c}{Wikitext}\\ 
      & \multicolumn{2}{c|}{LeNet5} &  \multicolumn{2}{c|}{Resnet18} & \multicolumn{2}{c|}{VGG16}   & \multicolumn{2}{c|}{ShuffleNet} & \multicolumn{1}{c|}{LSTM}  & \multicolumn{1}{c}{GPT-2} \\ \cline{2-11}
      & IID & Non-IID\tnote{2} & IID & Non-IID\tnote{2} & IID & Non-IID\tnote{2}& IID & Non-IID\tnote{2}& Non-IID\tnote{2} & Non-IID\tnote{2}\\
    \hline  
    FedAvg &   0.025    &  0.021     &    3.42   &   3.02    &  4.65   &  4.42 &   8.92    & 9.14    & 16.28    &    23.67    \\
     FedProx &  0.023 & 0.020    &  3.13   & 3.06     & 4.43  & 4.39       &   8.87  &  9.02      &  15.89   & 23.45 \\
    % \hline
    AFL &  0.019 &  0.019  &  2.69  &  2.66   & 4.02  & 4.11  & 8.31  &   8.54  &  14.12   &     21.89    \\
    TiFL &  0.018 &  0.016   &   2.31    &  2.01   &  3.56  & 3.31 & \underline{6.32}  &   6.73  &   13.25   &    20.56    \\
    Oort &  0.014 &   0.015    &   1.91  &   1.86   & 3.23   & 3.02 &  6.80 &    6.54    &  13.17    &   \underline{18.32}    \\ 
    % \hline
     Favor &  0.021 & 0.019   &  2.79   &  2.81    &  4.51  & 4.28  & 8.62  &   8.55  &  15.96  &    23.02     \\
    FedMarl &  \underline{0.012}    &   \underline{0.014}  &   \underline{1.82}  &  \underline{1.82}    &  \underline{3.17}    &  \underline{2.98}   &    6.72   &  \underline{6.29}   &   \underline{12.56}    &     19.26  \\
    \hline
    \model~ &  \textbf{0.007 }  &  \textbf{0.010}    &  \textbf{1.71}    &  \textbf{1.51  }  &  \textbf{ 2.26 }  & \textbf{2.71 }&  \textbf{5.90 } & \textbf{ 6.02}   &   \textbf{11.82}   &  \textbf{  17.85}     \\
    \bottomrule    
    \end{tabular}%
  \caption{The average energy consumption (KJ) of each training round for different selection approaches. \model~ achieves the best performance across all the datasets.}
  \label{tab:energy}%
    \begin{tablenotes}
    % \footnotesize 
    \scriptsize
    \item[1] Set $\beta =0.01$ for MNIST, $\beta =0.1$ for others to emulate Non-IID. Shakespeare and Wikitext are naturally non-IID.
    \end{tablenotes}
  \end{threeparttable}
    \vspace{-2mm}
\end{table*}%

\begin{figure*}[!ht]
    \centering
    \subfigure[MNIST (IID)]{\includegraphics[width=0.325\linewidth]{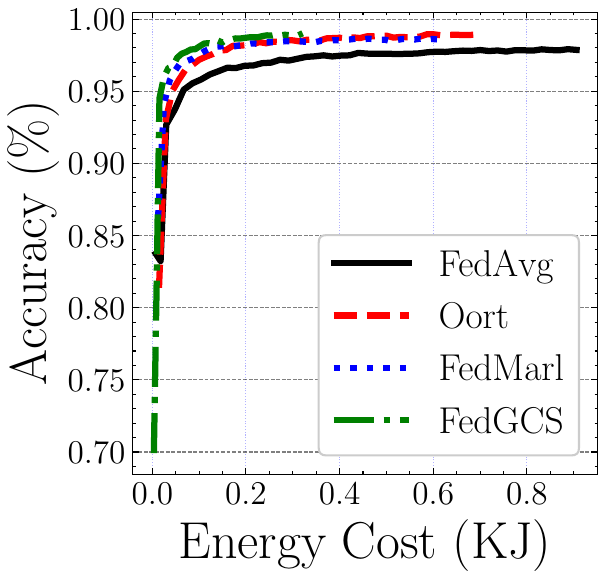}}
    \subfigure[MNIST (NonIID)]{\includegraphics[width=0.315\linewidth]{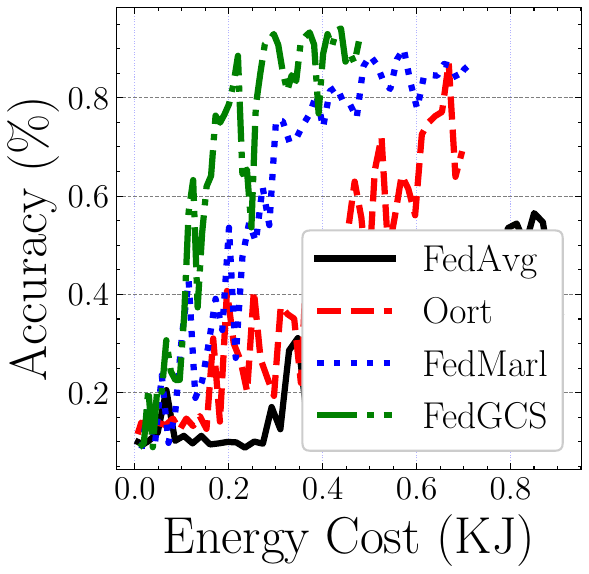}}
    \subfigure[CIFAR10 (IID)]{\includegraphics[width=0.315\linewidth]{Figures/energy.pdf}}
    \subfigure[CIFAR10 (NonIID)]{\includegraphics[width=0.315\linewidth]{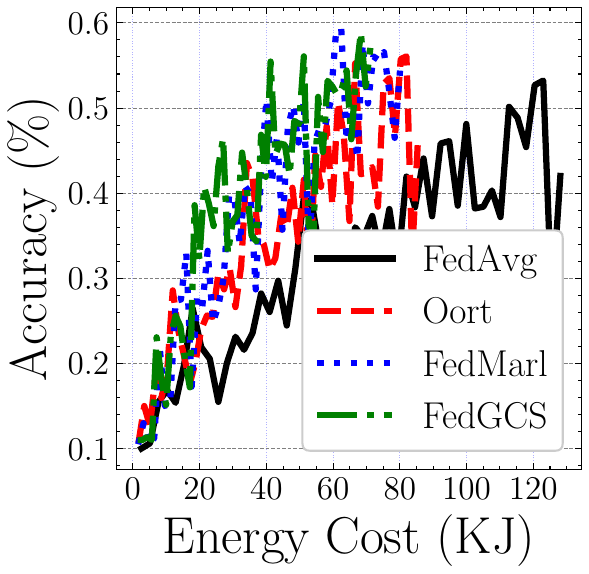}}
    \subfigure[CINIC10 (IID)]{\includegraphics[width=0.315\linewidth]{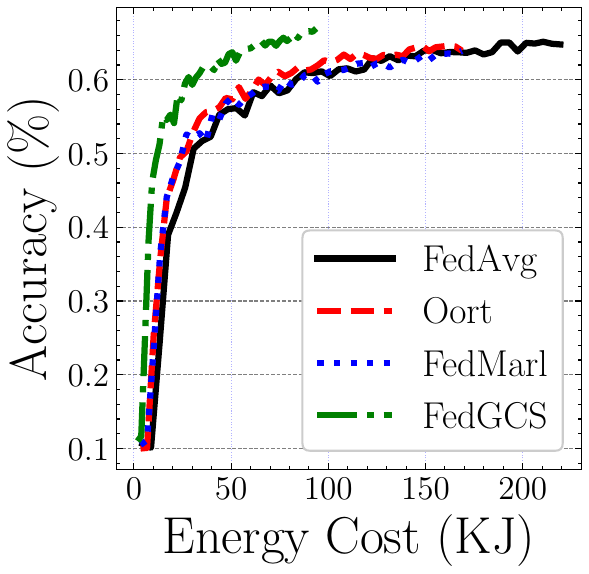}}
    \subfigure[CINIC10 (NonIID)]{\includegraphics[width=0.332\linewidth]{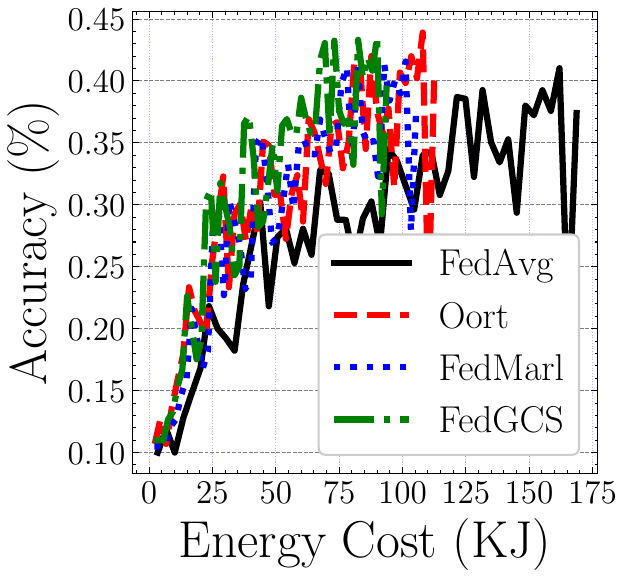}}
    \vspace{-4mm}
    \caption{ToE: time to energy. Efficiency comparison of various schemes.  
    }
    \label{fig:system-energy}
\end{figure*}

\item Learning-based Selection:
\begin{itemize}
\item \textbf{Favor}~\cite{Favor} utilizes accuracy to determine local weight selection, mitigating non-IID impacts and enhancing convergence. It involves training a DRL agent via a double-deep Q-learning Network (DQN). Despite evident disparities in local model weights, which hold guiding data for device selection, the DQN agent's training aims to optimize the expected total discounted reward, represented as \( R=\sum_{t=1}^T \gamma^{t-1} r_t=\sum_{t=1}^T \gamma^{t-1}\left(\Xi^{\left(\omega_t-\Omega\right)}-1\right) \).

\item \textbf{FedMarl}~\cite{fedmarl} applies multi-agent reinforcement learning to select devices for federated learning. It strategically selects RL agents each round to optimize accuracy, training latency, and communication cost. The reward $r_t$ for each round $t$ is given by:
\begin{equation}
r_t=w_1[U(A c c(t))-U(A c c(t-1))]-w_2 H_t-w_3 B_t
\end{equation}
where $H_t$, the processing latency, is:
\begin{equation}
H_t=\max _{1 \leq n \leq N}\left(H_{t, n}^p\right)+\max _{n: 1 \leq n \leq N, a_n^t=1}\left(H_{t, n}^{r e s t}+H_{t, n}^u\right)
\end{equation}
In this, $\max _{1 \leq n \leq N} H_{t, n}^p$ signifies the maximum time for generating probing losses, used by MARL agents for client selection and model update. The time for client $n$ to complete training and upload updates is $\max _{n: 1 \leq n \leq N, a_n^t=1}\left(H_{t, n}^{\text {rest }}+\right.$ $\left. H_{t, n}^u\right)$ Here, $U(.)$ is a utility function ensuring even modest improvements in $Acc(t)$ are recognized towards the end of the learning process, and $B_t$ represents the total communication cost.

\end{itemize}
\end{itemize}

\subsection{Implementation Details}
We ran Oor, Favor, and FedMarl 100 times each (for a total of 300 runs) to collect ``selection-score'' pair data to be used as training data for the subsequent model.
For the data augmentation part, we randomly shuffled each selection 25 times to model the order-independent properties of sets.
We adopt a single-layer LSTM as the Encoder and Decoder backbones and use two-layer feed-forward networks to implement
the Evaluator.
The hidden state sizes for the Encoder, Decoder, and Evaluator are 64, 64, and 200, respectively.
The embedding size of each device ID token is set to 32.
To train \model~, we set the batch size as 1024, the learning rate as 0.001, and $\alpha$ to 0.8. 
We utilize the top-25 device selection records as starting points for gradient-based optimization.
For FL part, we set the number of devices that are available in the device pool as $J=100$, while $T=10$ devices are selected to participate in each round and $T=20$ for FedMarl for further action-making. 
The number of training rounds $r$ is set to $r=10$ for GPT-2, and $r=50$ for others. 
The clients perform local training for $ep = 5$ epochs in each training round.

\section{Additional Results}

\subsection{Overall Performance of Latency}
Table~\ref{tab:latency} shows the total processing latency for all the methods.
Figure~\ref{fig:system-latency} shows the results of Time to Accuracy (ToA) in MNIST (IID), MNIST (Non-IID), CIFAR10 (IID), CIFAR10 (Non-IID), CINIC10 (IID), CINIC10 (Non-IID).

\subsection{Overall Performance of Energy Consumption}
Table~\ref{tab:energy} shows the total energy cost for all the methods.
Figure~\ref{fig:system-energy} shows the results of Energy to Accuracy (EoA) in MNIST (IID), MNIST (Non-IID), CIFAR10 (IID), CIFAR10 (Non-IID), CINIC10 (IID), CINIC10 (Non-IID).

\bibliographystyle{named}
\bibliography{8-ijcai}

% \bibliographystyle{named}
% \bibliography{8-ijcai}

\end{document}